\documentclass[pdftex,12pt]{article}
\usepackage[numbers]{natbib}
\pdfoutput=1  
\usepackage{arxiv}
\usepackage[utf8]{inputenc} 
\usepackage[T1]{fontenc}      
\usepackage{booktabs}       

\usepackage{microtype}      
\usepackage{graphicx}
\usepackage{bm}
\usepackage{amsmath,amsfonts,amssymb}
\usepackage{wrapfig}
\usepackage{subfigure}

\usepackage{multicol}
\usepackage[bookmarks=true]{hyperref}
\usepackage{hyperref}

\usepackage{wrapfig}
\usepackage{subfigure}

\usepackage{flexisym}
\usepackage{algpseudocode,algorithm}
\usepackage{lipsum}
\usepackage{enumerate}
\usepackage{wasysym}
\usepackage{soul}

\usepackage{algpseudocode}
\usepackage{tabularx}
\usepackage{hhline}
\usepackage{multirow}

\newlength\myindent
\title{Fully automatic structure from motion with a spline-based environment representation}
\author{
  Zhirui Wang\\
  Australian National  University\\
  \texttt{u5428281@anu.edu.cn} \\
   \And
 Laurent Kneip\\
  Shanghaitech  University\\
  \texttt{lkenip@shanghaitech.edu.cn} \\
}
\date{}

\begin{document}

\maketitle

\begin{abstract}
While the common environment representation in structure from motion is given by a sparse point cloud, the community has also investigated the use of lines to better enforce the inherent regularities in man-made surroundings. Following the potential of this idea, the present paper introduces a more flexible higher-order extension of points that provides a general model for structural edges in the environment, no matter if straight or curved. Our model relies on linked B\'ezier curves, the geometric intuition of which proves great benefits during parameter initialization and regularization. We present the first fully automatic pipeline that is able to generate spline-based representations without any human supervision. Besides a full graphical formulation of the problem, we introduce both geometric and photometric cues  as well as higher-level concepts such overall curve visibility and viewing angle restrictions to automatically manage the correspondences in the graph. Results prove that curve-based structure from motion with splines is able to outperform state-of-the-art sparse feature-based methods, as well as to model curved edges in the environment.
\end{abstract}

\section{Introduction}
\label{sec:introduction}

Reconstructing a 3D model from multiple planar projections is a classical \textit{inverse problem} with long-standing history in computer vision. The solution typically involves three steps. The first one is given by the extraction of stable features from the images, the second one by the establishment of correspondences between features in different images, and the third one by the exploitation of incidence relations that permit the recovery of camera poses and 3D structure. We denote a stable feature any point in the image that can be reliably extracted from different view-points while always pointing at the exact same point in 3D. Ignoring complex cases such as occlusions and apparent contours, these are all the image points for which a local extremum in the first order derivative can be observed. This---and the intuition behind line drawings---have lead to the initial belief that the most useful features in an image are simply all the edges. However, later research has shown that point correspondences are not only much easier to establish, but also easier to be used as part of incidence relations from which even closed-form solutions to camera resectioning and direct relative orientation can be derived \cite{nister2004efficient,hartley1997defense}. Point-based paradigms therefore are the dominating solution to the structure from motion problem.

A fundamental interest in edge-based structure from motion however remains. Edges provide more data, and thus must lead to higher accuracy. A common way to exploit the potential of edges while still enabling comfortable matching of features and exploitation of incidence relations is given by relying on straight lines. Many works have proven the feasibility of purely line-based structure from motion \cite{bartoli05}, or even hybrid architectures that rely simultaneously on points and lines \cite{pumarola17,zuo17,lu15}. The latter, in particular, have successfully demonstrated superior accuracy in comparison to purely point-based representations. The problem with lines is that they do not represent a general model for describing the 3D location of imaged edges; They are limited to specific, primarily man-made environments in which straight lines are abundant, either in the form of occlusion boundaries, or in the form of appearance boundaries in the texture.

More general approaches in which 3D information for curved edges is recovered have recently been demonstrated in the online, incremental visual localization and mapping community. Works such as \cite{engel2014lsd} and \cite{kuse2016robust} notably reconstruct semi-dense depth maps for all image edges. The depth estimates are updated and propagated from frame to frame. While certainly very successful in terms of an economic generation of enhanced map information, the representation is only local and therefore highly redundant in nature. A unique, global representation of the environment optimized jointly over all observations is not provided. Ignoring works that only focus on small scale object reconstruction \cite{delaunoy14}, the same accounts for fully dense approaches that estimate depth over the entire image \cite{newcombe2011dtam}.

Inspired by \cite{nurutdinova2015towards}, we present an incremental spline-based structure-from-motion pipeline that provides a unique, global and general higher-order model for edges in the environment (straight or bended). In particular, our contributions over the literature are:
\begin{itemize}
\item The first complete framework that automatically extracts, matches, initializes, and optimizes a purely curve-based environment representation without any human intervention. The optimization does not involve lifting, and hence remains fast on standard architectures.
\item Novel photometric and geometric criteria for verifying correspondences. In particular, our method uses color images, and the photometric error is evaluated in the HSV space. The quality of the correspondences is further reinforced by checking the consistency of edge identities and viewing directions.
\item A successful use of linked B\'ezier curves for representing 3D edges. The straightforward geometric meaning of B\'ezier spline parameters provides benefits during both initialization and regularization of the structure.
\end{itemize}
The paper is organized as follows. After a summary of further related work, Section \ref{sec:bezier} provides further details about the employed spatial parametrization. Section \ref{sec:automatic} then presents the core of our contribution, a fully automatic strategy for optimizing the curve-based representation. Section \ref{sec:experiments} finally concludes with experimental our experimental results, which confirm that structure from motion based on a high-order model is able to outperform point-based implementations.

\subsection{Further related work}

Curve-based geometric incidence relations have since ever intrigued the structure-from-motion community. For example, rather than solving the relative pose problem from points correspondences \cite{nister2004efficient,hartley1997defense}, early works such as \cite{porrill91} and lateron \cite{feldmar95} and \cite{kaminski04} looked into the possibility of using curves and surface tangents to solve the stereo calibration problem. However, the presented constraints for solving geometric computer vision problems are easily influenced by noise, and not practically useful. In order to improve the quality of curve-based structure from motion, further works therefore looked at special types of curves such as straight lines and cones, respectively \cite{faugeras95,kahl98}.

Our primary interest is the solution of structure-from-motion over many frames and observations. Point-based solutions are very mature from both theoretical \cite{hartley04} and practical perspectives \cite{agarwal09}. However, point-based representations are somewhat unsatisfying as they simply do not present a complete, visually appealing result. It is therefore natural that the structure-from-motion community has been striving for higher-level formulations that are able to return fully dense surface estimates \cite{delaunoy14}. Fully dense estimation is however very computationally demanding, which is why a compromise in the form of line based representations has also received significant attention from the community \cite{bartoli05,schindler06}. Lines are however not a general model for representing the environment, they fail in environments where the majority of edges are bended. \cite{kaess04,kahl03,xiao05,teney12} provide a solution to this problem by introducing curve-based representations of the environment, such as sub-division curves, non-rational B-splines, and implicit representations via 3D probability distributions. However, they do not exploit the edge measurements to improve on the quality of the pose estimations as well, as they do not optimize the curves and poses in a joint optimization process.

Full bundle adjustment over general curve models and camera poses has first been shown in \cite{berthilsson01}. The approach however suffers from a bias that occurs when the model is only partially observed. \cite{nurutdinova2015towards} discusses this problem in detail, and presents a lifting approach that transparently handles missing data. \cite{fabbri10} solves the problem by modeling curves as a set of shorter line segments, and \cite{cashman13} models the occlusions explicitly. While \cite{nurutdinova2015towards} is the most related to our approach, the lifted formulation is computationally demanding, the work does not discuss the fully automatic establishmentment of a correspondence graph that would enable fully automatic incremental structure-from-motion. Our main contribution is an efficient, fully automatic solution to this problem, thus enabling automatic curve-based structure from motion in larger scale environments.

Further related work can be found in the online visual SLAM community, which equally aimed at finding an efficient, general compromise between point-based \cite{klein2007parallel,mur2015orb} and dense \cite{newcombe2011dtam} formulations. While \cite{pumarola17,zuo17,lu15} have again looked at using lines (primarily through a combination with points), recent works such as \cite{engel2013semi,engel2014lsd,tarrio15,kuse2016robust} have also successfully realized visual SLAM pipelines based on general edge features by estimating depth along all edges that can identified in the images (e.g. by applying Canny-edge extraction \cite{canny1986computational}). While these methods do represent fully automatic pipelines, their results do rely on a global representation of curves, and consequently fail to jointly optimize over poses and structure.

\section{B\'ezier splines as a higher-order curve model}
\label{sec:bezier}

Different from the traditional map representation which uses points for map representation, we aim at using B\'ezier splines and thus reach a more complete, higher order representation of the environment. In this section, we are going to give a brief review of B\'ezier curves as well as the linked B\'ezier spline parametrization. We will conclude with an exposition of the basic registration cost for aligning a B\'ezier curve with a set of pixels in an image.

\subsection{A short review of B\'ezier splines}

A B\'ezier-spline is a continuous curve expression parametrized as a function of control points and a continuous curve parameter $t$. Every point on the curve can be obtained by referring to a unique value of $t$. The general definition of a B\'ezier spline is:
\begin{equation}
\mathbf{B}(t)  = \sum_{i=0}^{k}b_{i,k}\mathbf{P}_i,
\label{eq:bezier}
\end{equation}
where $b_{i,k} = \begin{pmatrix} k\\i \end{pmatrix}t^i(1-t)^{k-i}$ and $\begin{pmatrix} k\\i \end{pmatrix} = \frac{k!}{i!(k-i)!}$. $\mathbf{P_i}$ is the $\mathbf{i}^{th}$ control point of the B\'ezier curve, and $k$ is the degree of the curve. In our work, we use cubic B\'ezier splines for representing curves as they are a powerful representation allowing for independent spatial gradients at the beginning and the end of the curve. Cubic splines employ four control points $\mathbf{P_i}$, hence $k=3$, and
\begin{equation}
\mathbf{B}(t)  = (1-t)^3 \mathbf{P}_0 + 3(1-t)^2 t \mathbf{P}_1 + 3(1-t) t^2 \mathbf{P}_2 + t^3 \mathbf{P}_3.
\label{eq:bezier_curve}
\end{equation}
Besides compactness, the choice of B\'ezier splines is motivated by their simple geometry meaning: the first and last control points are simply the beginning and ending point of the curve while the directions from the first control point to the second and the fourth control point to the third are equal to the local gradient at the beginning and the end of the curve. As we will see, this clear geometric meaning facilitates initialization and regularization of parameters. Furthermore B\'ezier curves are invariant with respect to affine transformations, which means that the transformation of a B\'ezier curve between different coordinate frames is done by simply applying the same transformation to its control points. Furthermore, B\'ezier splines provide implicit smoothness and a scale invariant density of points along the curve by simply adjusting the step-size of $t$.

\subsection{Smooth polyb\'eziers}

A single cubic B\'ezier spline is unable to fit arbitrarily complex contours. We overcome this problem by borrowing a simple idea from computer graphics: composite, piece-wise B\'ezier curves (i.e. so-called polyb\'eziers). As indicated in Figure \ref{fig:bezier_init_a}, polyb\'eziers separate a contour into multiple segments where every segment is represented by a single B\'ezier spline of limited order. Composite splines stay continuous by simply sharing the ending point of a segment with the starting point of the subsequent segment. Moreover, in order to maintain smoothness, we furthermore make sure that the gradient at the end of one segment coincides with the spatial gradient at the beginning of the next.

Imposing these constraints is simply done by making the control points a function of other latent variables that are shared among neighbouring segments. On one hand, the continuity simply requires the first control point of one segment to be equal to the last control point of the previous segment. Let $\mathbf{P}_{0}^{i} \ldots \mathbf{P}_{3}^{i}$ be the control points of segment $i$. Furthermore, let the first control point of a segment also be denoted by $\mathbf{Q}_{i}$. We obtain $\mathbf{P}_1^i = \mathbf{Q}_{i}$ and $\mathbf{P}_3^i = \mathbf{Q}_{i+1}$. On the other hand, sharing the gradient is made possible by explicitly introducing the local direction at the beginning of each segment, denoted $\mathbf{v}_{i}$. $\mathbf{v}_{i}$ is a 3-vector constrained to unit-norm, which means it is a spatial direction with two degrees of freedom only. Since the second control point by definition lies in the direction of the local gradient at the first control point, it may be parametrised as $\mathbf{P}_2^i = \mathbf{Q}_{i} + d_1^i\mathbf{v}_{i}$. In order to guarantee smoothness, the third control point of segment $i$ in turn becomes a function of the gradient at the beginning of the next segment, i.e. $\mathbf{P}_3^i = \mathbf{Q}_{i+1} - d_2^i\mathbf{v}_{i+1}$.

In summary, an arbitrary curve is given by a sequence of parameters $\mathbf{Q}_{i}$, $\mathbf{v}_{i}$, and $\{d_1^i,d_2^i\}$.  By sharing parameters between neighboring segments, a 3D composite B\'ezier curve is represented as a sequence of cubic B\'ezier segments where
\begin{equation}
\mathbf{P}_1^i = \mathbf{Q}_{i}\text{, }
\mathbf{P}_2^i = \mathbf{Q}_{i} + d_1^i\mathbf{v}_{i}\text{, } \\
\mathbf{P}_3^i = \mathbf{Q}_{i+1} - d_2^i\mathbf{v}_{i+1}\text{, } \\
\mathbf{P}_4^i = \mathbf{Q}_{i+1}.
\label{eq:controlPoints}
\end{equation}
We parametrize the normal vectors $\mathbf{v}_{i}$ minimally by making them a function of two rotation angles $\bm{\theta}_{i}$, which avoids the addition of side-constraints during optimization. However, in order to facilitate optimisation and avoid the gimbal lock, the parametrisation is local about an initial direction expressed by a pre-rotation $\mathbf{R}_{0}$, i.e. $\mathbf{v}_{i} = \mathbf{R}_{0} \mathbf{v}(\bm{\theta}_{i})$.

\subsection{Initialization of B\'ezier splines}
\label{subsec:spline_init}

\begin{figure}[t]
\centering
  \subfigure[\label{fig:bezier_init_a}]{\includegraphics[width=0.45\textwidth]{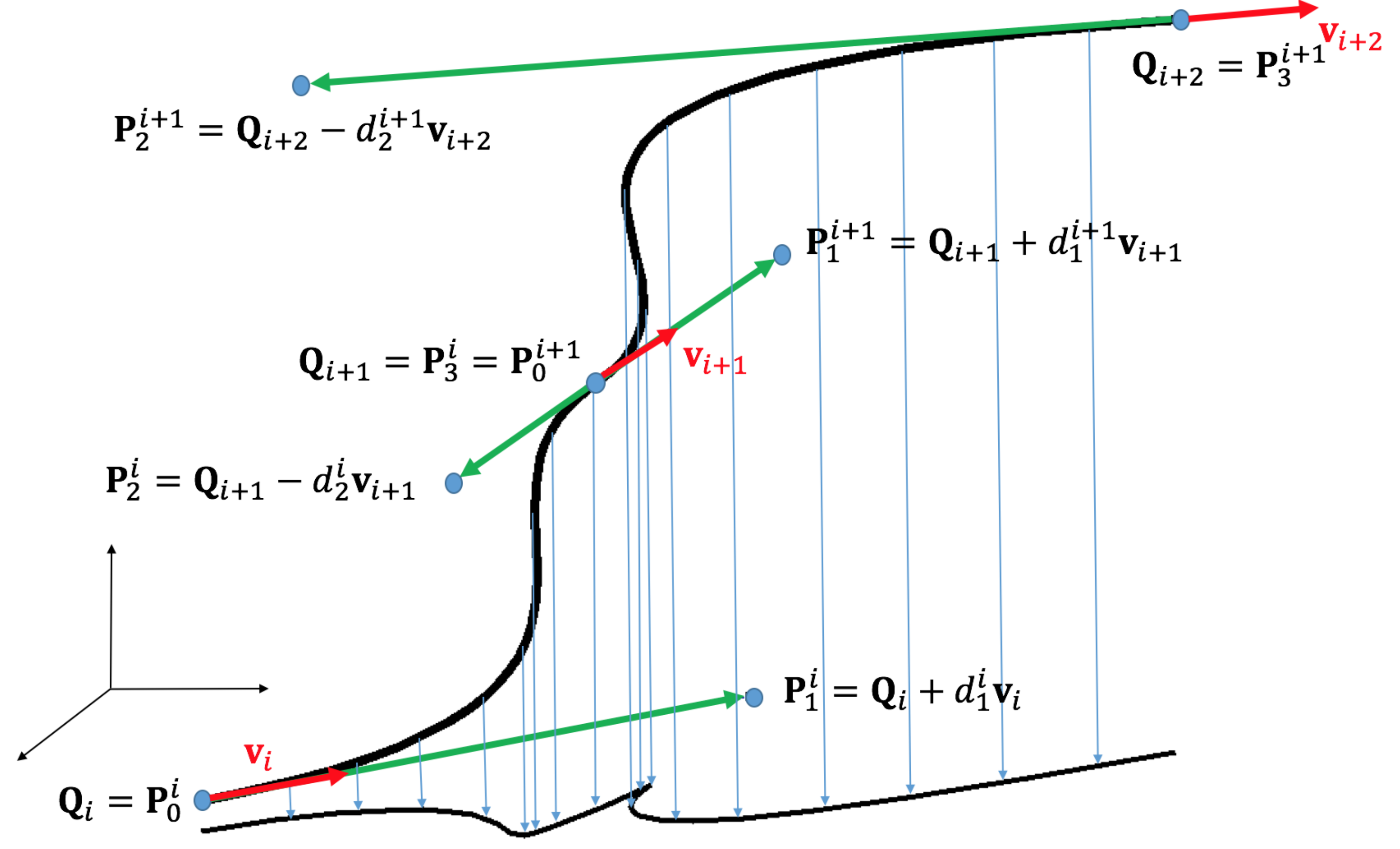}}
  \subfigure[\label{fig:bezier_init_b}]{\includegraphics[width=0.45\textwidth]{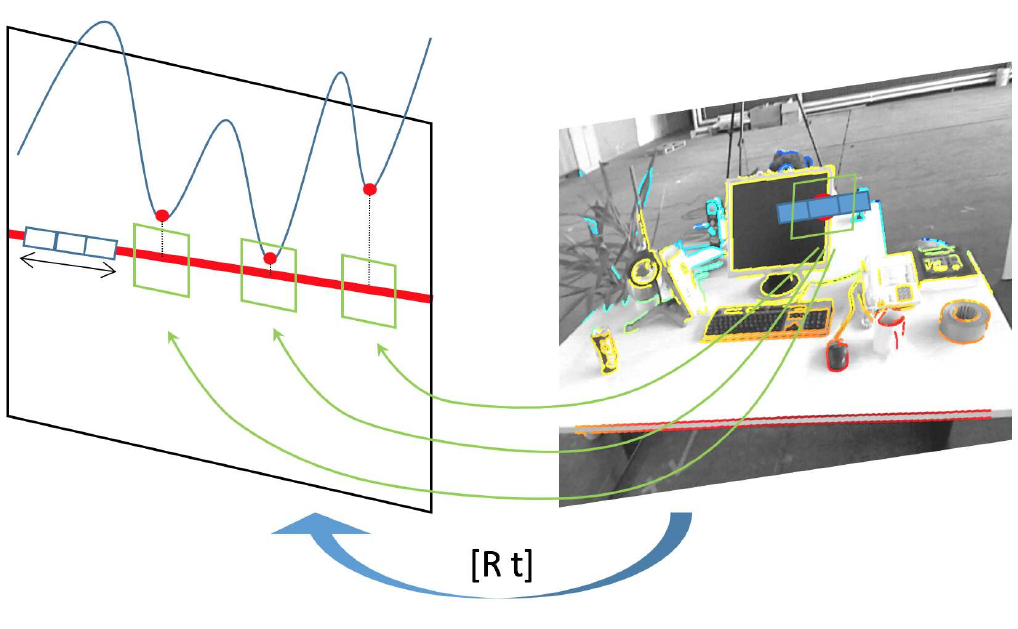}}
  \caption{Left: Two segments of a smooth 3D b\'ezier curve. Some of the optimization variables (the control points $\mathbf{Q}_{i}$ and the gradient directions $\mathbf{v}_{i}$) are shared among adjacent segments. Right: Epipolar matching with 1D patches. The three best local minima are sub-sequently disambiguated by 2D patch matching. The colors in the right frame indicate the depth of pixels and thus show an example result of semi-dense matching.}
  \label{fig:bezier_init}
\end{figure}
As we will see in Section \ref{sec:automatic}, our method relies on a sparse technique to first initialize the poses of all frames in a video sequence. We proceed by extracting Canny-edges \cite{canny1986computational} in each image and grouping them into curves based on simple connectivity and thresholding of local curvature. We then initialize the depth of each pixel on an edge by using a variant of the semi-dense epipolar tracking method presented in \cite{engel2013semi}. For each pixel within each group, we perform the following steps to recover the depth:
\begin{itemize}
\item Find a good reference frame for stereo matching by considering the length of the baseline and the parallelism between the epipolar direction and the local image gradient.
\item Extract the epipolar line in the reference frame.
\item Perform a 1D search for photometric consistency along the epipolar line by comparing 1D image patches.
\item Short-list the three best local minima.
\item For each local minimum, perform a 2D patch comparison to find the best.
\item Minimise the photometric error via sub-pixel refinement of the disparity along the epipolar line.
\item Take a robust average among all the depths recovered within a local window.
\end{itemize}
The procedure is visualised in Figure \ref{fig:bezier_init_b}. Knowing camera poses as well as semi-dense depth maps in each frame, it now becomes possible to initialize the 3D position of points along the curve expressed in world coordinates $\mathbf{p}_{w}$
\begin{equation}
\mathbf{p}_{w} = \mathbf{R}_{w} (d \mathbf{K}^{-1} \mathbf{p}) + \mathbf{t}_{w}
\label{eq:transform}
\end{equation}
where  $\mathbf{R}_{w} \text{ and }\mathbf{t}_{w}$ are the rotation and translation transforming a point $\mathbf{p}$ from the camera to the world coordinate frame, $\mathbf{K}$ is the matrix of intrinsic camera parameters, and $d$ if the depth of the point along the principal axis. We transform all points for which a depth has been recovered into 3D, and furthermore separate the contours into segments such that each segment has roughly the same number of pixels. We furthermore assume that each segment can now be correctly represented by a cubic B\'ezier spline. We finally estimate the control points of each segment by counting the number of pixels composing the segment, sampling an equal number of values for $t$ distributed homogeneously within the interval $(0,1)$, and assuming that the B\'ezier spline evaluated at those continuous curve parameter values leads exactly to the hypothesised world point. Let us assume that there are $m$ world points. Under the above assumption, the curve parameter for each one of the points becomes
\begin{equation}
  t_{k} = \frac{k}{m-1} , k \in \{0, \ldots, m-1\}
\end{equation}
Using (\ref{eq:bezier_curve}), we can then find the control points by constructing and solving the linear problem
\small
\begin{equation}
  \left(\begin{matrix} (1-t_0)^3 \mathbf{I} & 3(1-t_0)^2 t \mathbf{I} & 3(1-t_0) t^2 \mathbf{I} & t_0^3 \mathbf{I} \\
                                \ldots & \ldots & \ldots & \ldots \\
                                (1-t_{m-1})^3 \mathbf{I} & 3(1-t_{m-1})^2 t \mathbf{I} & 3(1-t_{m-1}) t^2 \mathbf{I} & t_{m-1}^3 \mathbf{I}
  \end{matrix}\right)
  \left(\begin{matrix}
	  \mathbf{P}_0 \\
	  \mathbf{P}_1 \\
	  \mathbf{P}_3 \\
	  \mathbf{P}_4
  \end{matrix}\right) = 
  \left(\begin{matrix}
	  \mathbf{p}_{w,0} \\
	  \ldots \\
	  \mathbf{p}_{w,m-1}
  \end{matrix}\right)
\end{equation}
\normalsize
Note that the left-hand matrix only depends on a discrete number of homogeneously sampled values for $t$, and therefore can be computed upfront. To conclude the initialisation, the B\'ezier segments are grouped into curves following the same order then the original segments extracted in the image. To enforce continuity and smoothness in the initialised curve, the first control point of each segment is simply replaced by the last control point of the previous segment, and the direction at the link point and the distance from the link point are set to
\begin{eqnarray}
  \mathbf{v}_{i} & = & 0.5 * (\mathbf{P}_4^{(i-1)}-\mathbf{P}_3^{(i-1)}) / \| \mathbf{P}_4^{(i-1)}-\mathbf{P}_3^{(i-1)} \| \nonumber\\ & + & 0.5 * (\mathbf{P}_2^i-\mathbf{P}_1^i) / \| \mathbf{P}_2^i-\mathbf{P}_1^i \| \nonumber \\
  d_1^i & = & \| \mathbf{P}_2^i-\mathbf{P}_1^i \| \nonumber \\
  d_2^i & = & \| \mathbf{P}_4^i-\mathbf{P}_3^i \| \nonumber.
\end{eqnarray}

\section{Fully automatic, spline-based structure from motion}
\label{sec:automatic}

This section explains the structure of our curve-based structure-from-motion problem, which can notably be formulated as a graph optimisation problem. The first part of the section assumes that the structure of this graph is already initialised, and in turn focusses on how the individual registration costs as well as the overall bundle adjustment are computed. The second part of the section then presents the detailed flow-chart of our incremental structure-from-motion, and in particular provides all the details on the automatic management of the graph structure (i.e. correspondences between segments and frames).

\subsection{Curve-based structure-from-motion as a graphical optimisation problem}

\begin{figure*}[t]
\centering
\includegraphics[width=\textwidth]{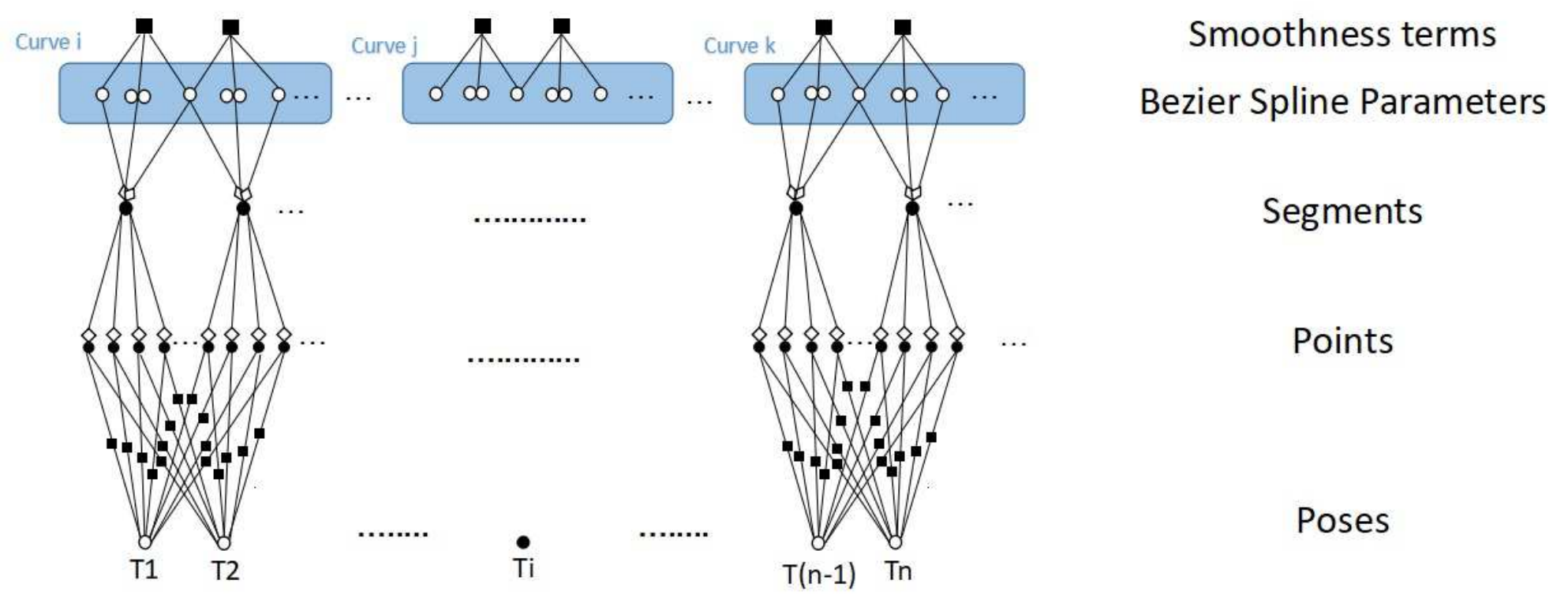}
\caption{Factor graph of our optimisation problem. Curves are composed of B\'ezier segments, which in turn are sampled to return 3D points. The latter are reprojected into frames if a correspondence between this segment and that frame exists. The curve parameters are not directly optimised, but depend on latent variables some of which are shared among neighbouring segments (bordering control points, as well as curve directions in those points).}
\label{fig:graph}
\end{figure*}

The overall factor graph of our optimisation problem is illustrated in \ref{fig:graph}. Our map is given as a set of composite B\'ezier curves, each one being composed of one or multiple cubic B\'ezier splines, which we call here \textit{segments}. However, as explained in the previous section, in order to ensure continuity and smoothness in the curves, we do not optimise the control points of the segments directly. The control points of the segments are dependent on latent variables, which potentially are even shared across neighbouring segments. With respect to Figure \ref{fig:graph}, these parameters are simply called \textit{B\'ezier spline parameters}. In order to prevent the latter from collapsing or drifting off into unobservable directions, we have our first cost terms added to the graph, which are regularisation constraints on the B\'ezier spline parameters. We then sample points from each segment, and reproject them into frames for which a correspondence has been established. A second type of cost term occurs here in the form of our 3D-to-2D curve registration loss. The correspondence management will be discussed later, here we focus on the cost terms and the actual optimisation of the graph.

Let us define the vector $\mathbf{b}_{i} = \left[ \begin{matrix} \mathbf{Q}_{i}^{T} & \mathbf{Q}_{i+1}^{T} & \bm{\theta}_i^{T} & \bm{\theta}_{i+1}^{T} & d_1^i & d_2^i \end{matrix} \right]^{T}$ as the vector of parameters defining the B\'ezier spline $\mathbf{B}_{i}$, which may hence be written as a function $\mathbf{B}(\mathbf{b}_i,t)$. It is clear that many of the parameters are shared among different segments, but we ignore this here for the sake of a simplified notation. We assume to have $n$ splines. Let us furthermore assume that we have $m$ camera poses and that the pose of each camera is parametrized by the 6-vector $\delta\pi_j$ that expresses a local change with respect to the original pose $\pi_{j,0}$. Let $f_{\delta\pi_j + \pi_{j,0}}(\cdot)$ be a function that transforms a point from the world frame into the image plane of a camera at position $\delta\pi_j + \pi_{j,0}$. The function $f$ assumes and uses known camera intrinsic parameters. Let $\mathbf{C}^{j}$ denote all pixels along an edge in keyframe $j$, and $\eta( \mathbf{x}, \mathbf{C}^{j} )$ a function that returns the nearest pixel on an edge (i.e. within $\mathbf{C}^{j}$) to a reprojected image location $\mathbf{x}$. The final objective of the global optimisation is given by
\begin{equation}
  \left\{ \delta\hat{\pi}_{1}, \cdots, \delta\hat{\pi}_{m}, \hat{\mathbf{b}}_{1}, \cdots, \hat{\mathbf{b}}_{n} \right\} = \underset{\delta\pi_{1}, \cdots, \delta\pi_{m}, \mathbf{b}_{1}, \cdots, \mathbf{b}_{n}}{\operatorname{argmin}} E_{\text{geo}} + \lambda_{1} E_{s1} + \lambda_{2} E_{s2} \label{SDBA},
\end{equation}
where
\begin{eqnarray}
 E_{\text{geo}}  & = & \sum_{i=1}^n \sum_{j=1}^m \mathbf{1}_{ij} \sum_{k=0}^{s} \mu \left( \left( \mathbf{x}_{ijk} - \eta( \mathbf{x}_{ijk}, \mathbf{C}^{j} ) \right)^{T} \cdot g(  \eta( \mathbf{x}_{ijk}, \mathbf{C}^{j} ) ) \right)\nonumber \\
 \mathbf{x}_{ijk} & = & f_{\delta\pi_j + \pi_{j,0}}\left( \mathbf{B}(\mathbf{b}_i,t_{k})  \right) \nonumber \\
 E_{s1} & = & \sum_{i=1}^n ( \| \mathbf{Q}_{i} - \mathbf{Q}_{i+1} \| - \Delta\mathbf{Q}_{i,0} )^{2} \nonumber \\
 E_{s2} & = & \sum_{i=1}^n \left( \measuredangle^{2}( \mathbf{Q}_{i+1} - \mathbf{Q}_{i}, \mathbf{v}(\bm{\theta}_{i}) ) + \measuredangle^{2}( \mathbf{Q}_{i+1} - \mathbf{Q}_{i}, \mathbf{v}(\bm{\theta}_{i+1}) ) \right) \nonumber .
\end{eqnarray}
$\mathbf{1}_{ij}$ is an indicator function that equals to $1$ if the segment $i$ is visible in frame $j$, and otherwise to $0$. Index $k$ runs from $0$ to $s$ causing a homogeneous sampling of 3D points along the segment $i$ through the continuous curve parameters $t_{k}$. $\mathbf{x}_{ijk}$ results as the $k$-th 3D point sampled along the spline $i$ and reprojected into the view $j$. $E_{\text{geo}}$ as a result denotes the geometric registration cost given as the sum of disparities between reprojected 3D points $\mathbf{x}_{ijk}$ and their nearest neighbours in $\mathbf{C}^{j}$. A projection onto the local gradient direction $g(  \eta( \mathbf{x}_{ijk}, \mathbf{C}^{j} ) )$ is added in order to facilitate optimisation in the \textit{sliding situation}. Note that, as discussed in \cite{zhouyi_icra17}, this step also helps to efficiently overcome the bias discussed in \cite{nurutdinova2015towards} without having to employ the more expensive technique of variable lifting. To conclude, a robust norm $\mu(\cdot)$ such as the Huber norm is added to account for outliers and missing data.

Besides the residuals in the form of curve alignment errors, we additionally have the regularisation costs $E_{1}$ and $E_{2}$, which are weighted in using the trade-off parameters $\lambda_{1}$ and $\lambda_{2}$. The regularization terms are added to the cost function in order to prevent convergence into wrong local minima. $E_{1}$ enforces the length of each segment (i.e. the distance between the first and the last control point) to be consistent with its original value after initialization. The term introduces a penalty in the situation where the algorithm aims at collapsing a segment such that the entire segment would match to a single pixel, and thus return a very low registration cost. Second, because curves are composed of relatively short segments, we want to prevent single segments from presenting too high curvature. $E_{2}$ penalises high curvature by making sure that the spatial curve directions in the first and the last control point (indicated by $\mathbf{v}(\bm{\theta}_{i})$ and $\mathbf{v}(\bm{\theta}_{i+1})$) do not deviate too much from the vector between points $\mathbf{Q}_{i}$ and $\mathbf{Q}_{i+1}$. $E_{2}$ is particularly helpful to prevent uncontrolled curve behaviours in a situation where dept his badly observable.

Initial poses are given from a sparse initialisation, and the initial values for the B\'ezier splines are obtained using the procedure outlined in Section \ref{subsec:spline_init} (further details about graph management and initialisation are given in the following section). After initialisation, we solve the bundle adjustment problem (\ref{SDBA}) using an off-the-shelf nonlinear implementation of Levenberg-Marquardt \cite{ceres-solver}. The latter iteratively performs local linearizations of the residual and regularisation terms in order to gradually update the pose and B\'ezier spline parameters.

\subsection{Efficient residual error computation}

\begin{wrapfigure}{r}{0.3\textwidth}
  \vspace{-1.1cm}
  \centering
  \includegraphics[width=0.2\textwidth]{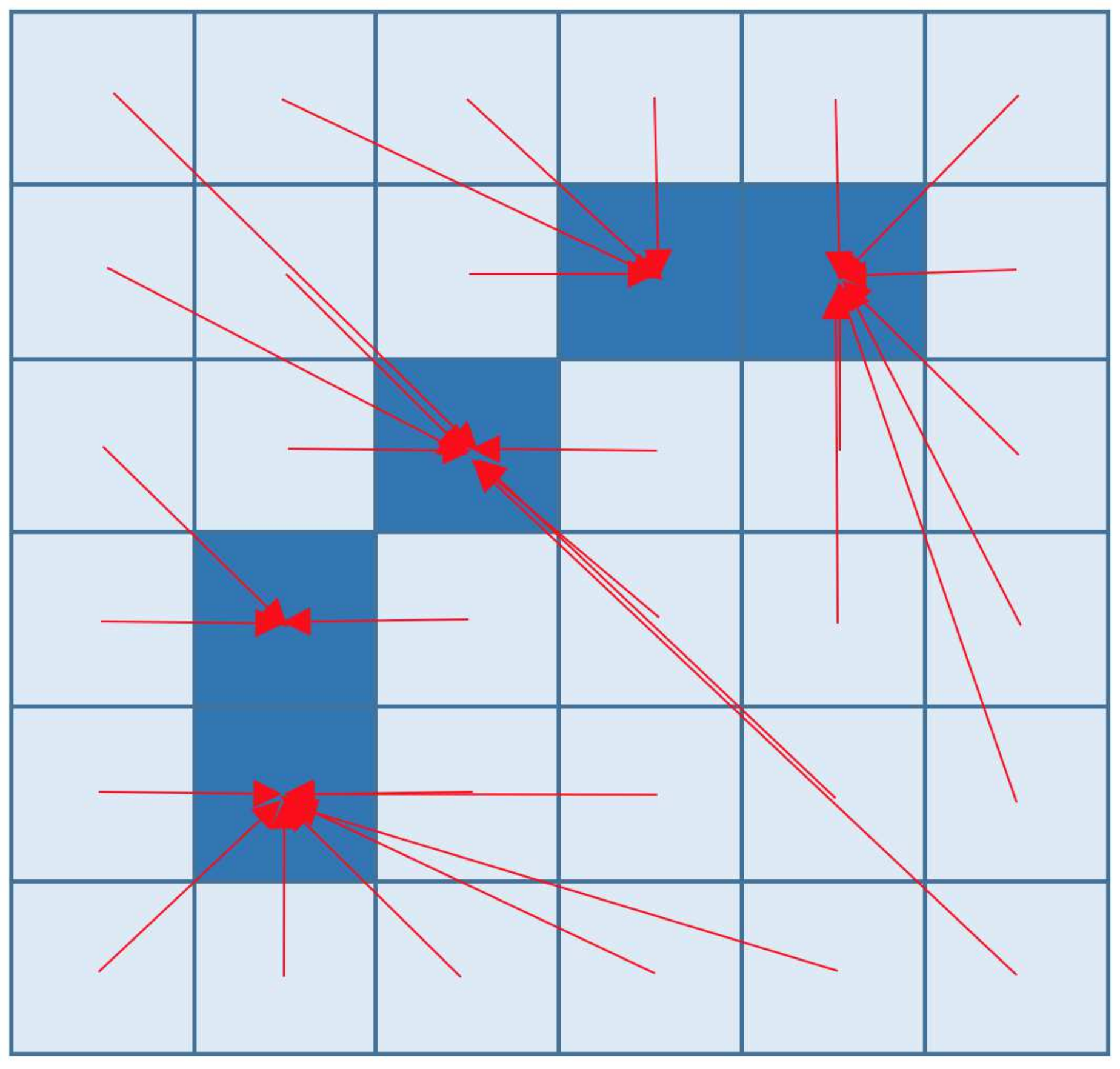}
  \caption{Example nearest neighbour field.}
  \vspace{-1cm}
  \label{fig:splineInitialization}
\end{wrapfigure}
One of the more expensive parts of the computation is given by the nearest neighbour look-up $\eta(\mathbf{x},\mathbf{C}^{j})$. Inspired by \cite{zhouyi_icra17}, we employ a simple solution to speed up the optimisation by pre-computing a nearest neighbour look-up field that indicates the nearest pixel on an edge for any pixel in the entire image. An example is given in Figure \ref{fig:splineInitialization}. The extraction of the nearest neighbour field is accelerated by limiting it to pixels which are at most 15 pixels away from an edge.

\subsection{Overall flow-chart}

\begin{figure}[t!]
\centering
\includegraphics[width=0.90\textwidth]{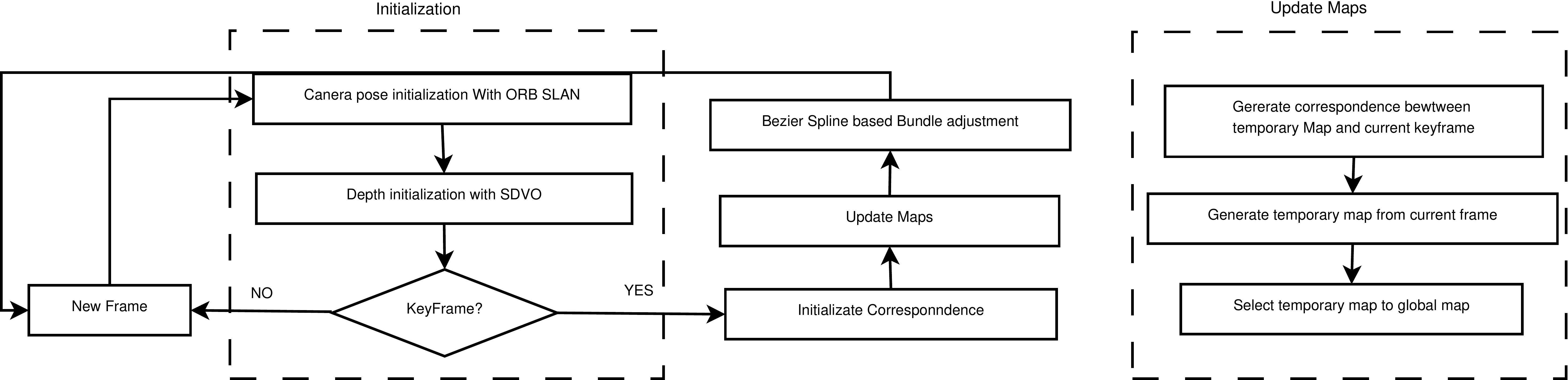}
\caption{Overall flow-chart of our B\'ezier spline-based structure from motion framework including the initialisation from a sparse point-based method.}
\label{fig:Pipeline}
\end{figure}

The input of the system is simply a sequence of RGB image from a calibrated camera. Before we initialize the B\'ezier Map, we perform ORB SLAM \cite{mur2015orb} to obtain an initial guess for the camera positions. ORB SLAM is a sparse feature based simultaneous localization and mapping system which can provide an accurate guess of the camera position in real-time. With the initial camera positions in hand, we then incrementally parse our frames and initialise new B\'ezier splines. Each time a new keyframe is added, we first establish the correspondence with existing segments before adding potentially new segments. The initialisation of splines from a single frame uses the strategy proposed in Section \ref{subsec:spline_init}. A flow-chart of the overall system is indicated in Figure \ref{fig:Pipeline}.

The B\'ezier splines are grouped into two distinct maps, one global map that stores well observed splines and a temporary map that stores new spline initialisations. All B\'ezier splines are initially put into the temporary map and then moved to the global map once sufficient observations are available. This delayed initialisation scheme helps to robustify the optimisation, as bundle adjustment uses only the well-observed splines in the global map. In order to prevent the addition of redundant representations, the establishment of correspondences in new keyframes (outlined in Section \ref{sec:correspondences}) needs to first consider the segments in the global map before moving on to the temporary map.

New segments are added to the temporary map whenever a sufficiently large group of connected pixels has not been registered with existing splines, and the semi-dense depth measurement for those pixels succeeded. The segment is added to an existing curve if the seed group of pixels is smoothly connected to the pixels of an already existing curve. To prevent the algorithm from losing too many correspondences in difficult passages, newly initialized B\'ezier splines may also be added directly into the global map to keep the tracking of subsequent frames alive. The addition of a new keyframe is concluded by local bundle adjustment over all recently observed frames and landmarks in the global map. Segments with less than three observations in keyframes will not be considered for updating the pose of the cameras. We alternately fix the parameters of B\'ezier splines and camera poses and optimize the other. After all key-frames have been loaded, we perform global bundle adjustment over all frames and splines.

\subsection{Correspondence establishment}
\label{sec:correspondences}

In this section, we are going to explain how we establish and manage the correspondences between segments and key-frames. Correspondences are verified based on four criteria:
\begin{itemize}
\item Spatial distance: We require the initial geometric registration cost to be small enough. Points from a spline are required to consistently reproject near a set of connected edge pixels in the image. We evaluate the average reprojection error. Only splines with an average error lower than a given threshold will be considered as an inlier correspondence.
\item Appearance-based error: We store the average color of a segment in its original observation. We do not only consider the pixels forming the edge itself, but an isotropically enlarged region around each segment. We again set a threshold on the difference in appearance for determining inliers.
\item Viewing direction: Since the appearance in the neighbourhood of edges can depend heavily on the viewing direction (in particular for occlusion boundaries), we add a limitation on the range of possible viewing directions. We assume the viewing direction of an observation to be the vector from the camera center to center of a segment transformed into the world frame. A correspondence is no longer established if the current viewing direction has an angle of more than sixty degrees away from the average viewing direction of all previous observations.
\item Depth check: We check the relative depth of each reprojected segment, and discard segments with negative depth.
\end{itemize} 
We further add correspondence pruning based on weak overall curve observations. For curves where more than fifty percent of all segments have no longer been observed in the three most recently added keyframes, the entire curve will be disabled. A disabled curve will no longer be used in local bundle adjustment until it is reactivated by sufficiently new observations in a new frame.

The average color error between a segment in its original image and the closest pixel retrieved via the nearest neighbor field is evaluated as follows. We represent the color in HSV format where, thus returning the average hue $H$, saturation $S$, and lightness $L$ values. Operating in the HSV color space can be more robust to illumination changes and difference caused by viewing direction changes. However, due to its definition, the hue value becomes unobservable at zero saturation. Errors in H therefore need to be scaled by the average saturation. The error between two segments is finally given by
\begin{equation}
\mathbf{E} = \lambda * \frac{S_1 + S_2}{2} * \textbf{abs}(H_1 - H_2) + (L_1 - L_2)
\end{equation}

\section{Experimental evaluation}
\label{sec:experiments}

The algorithm is implemented in C++ and depends on OpenCV and the Google Ceres-solver \cite{ceres-solver} for solving our curve-based bundle-adjustment. We evaluate the pipeline on several indoor and outdoor images from open-source benchmark sequences \cite{sturm2012benchmark,handa:etal:ICRA2014,Geiger2012CVPR}. Although some of them are RGB-D datasets, we only use the RGB channel in our pipeline. For initialization, we precompute the camera poses using ORB-SLAM \cite{mur2015orb}, and also compare our solution against its global optimization result. We evaluate our result in terms of both the accuracy of the camera trajectory and the quality of the environment mapping.

\subsection{Evaluation on Simulation Datasets}

Before we test our pipeline on a larger scale dataset, we perform a dedicated experiment where the performance of edge-based registration is analysed and compared against a state-of-the-art point-based solution in different environments. Each test is generated by taking an image of the real-world, assuming it to be a planar environment, and then generating novel views by assuming a circular orbit on top of the plane (new views can easily be generated by homography warping). This allows us to work with features from the real world, but at the same time focuses the experiment on the actual accuracy of the estimation (i.e. issues related to the correspondence establishment are not taken into account). It furthermore allows us to explore a larger variety of environments while each time maintaining information about the ground-truth trajectory.

\begin{table}[h]
\label{fig:table1}
\begin{tabular}{p{0.2\linewidth}p{0.12\linewidth}p{0.12\linewidth}p{0.12\linewidth}p{0.12\linewidth}p{0.12\linewidth}|}
\hline
Noise & $0\%$ & $5\%$  & $10\%$ & $15\%$ & $20\%$ \\
\hline
Logos (m) & 0.016215 & X & X & X & X\\
Indoor imgs (m) & 0.000703 & 0.000738 & 0.000691  & X & X \\
Outdoor imgs (m) & 0.001634 &  0.000962 & X & X & X \\
\hline
\end{tabular}
\caption[]{Average Position Error of ORB SLAM on different types of images ('/' means failure)}
\end{table}

\begin{table}[h]
\label{fig:table2}
\begin{tabular}{p{0.2\linewidth}p{0.12\linewidth}p{0.12\linewidth}p{0.12\linewidth}p{0.12\linewidth} p{0.12\linewidth}|}
\hline
Noise & $0\%$ & $5\%$  & $10\%$ & $15\%$ & $20\%$ \\
\hline
Logos (m) & 0.000415 & 0.000312 & 0.000458 & 0.0035 & 0.0111\\
Indoor imgs (m) & 0.000533 & 0.000533 & 0.000459  & 0.0014 & 0.0046 \\
Outdoor imgs (m) & 0.000885 &  0.0015 & 0.00153 &  0.0042 & 0.0068 \\
Overall (m) & 0.000611 & 0.000798 & 0.000812 & 0.003 & 0.0075\\
\hline
\end{tabular}
\caption[]{Average Position Error of our method on different types of images.}
\end{table}

\begin{figure}[b!]
\centering
\subfigure{\includegraphics[width=0.3\textwidth]
{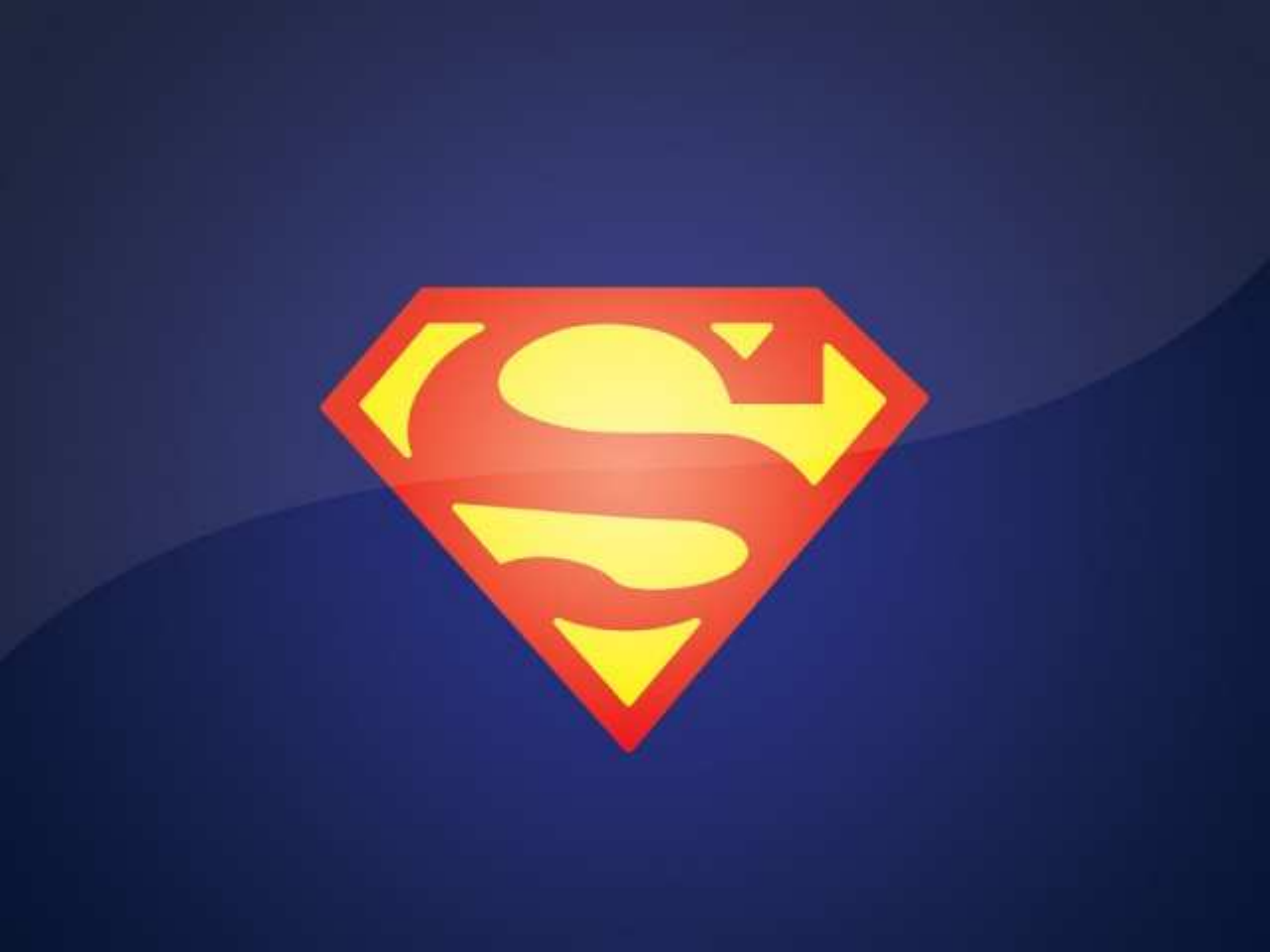}\label{exp:f1}}
\subfigure{\includegraphics[width=0.3\textwidth]
{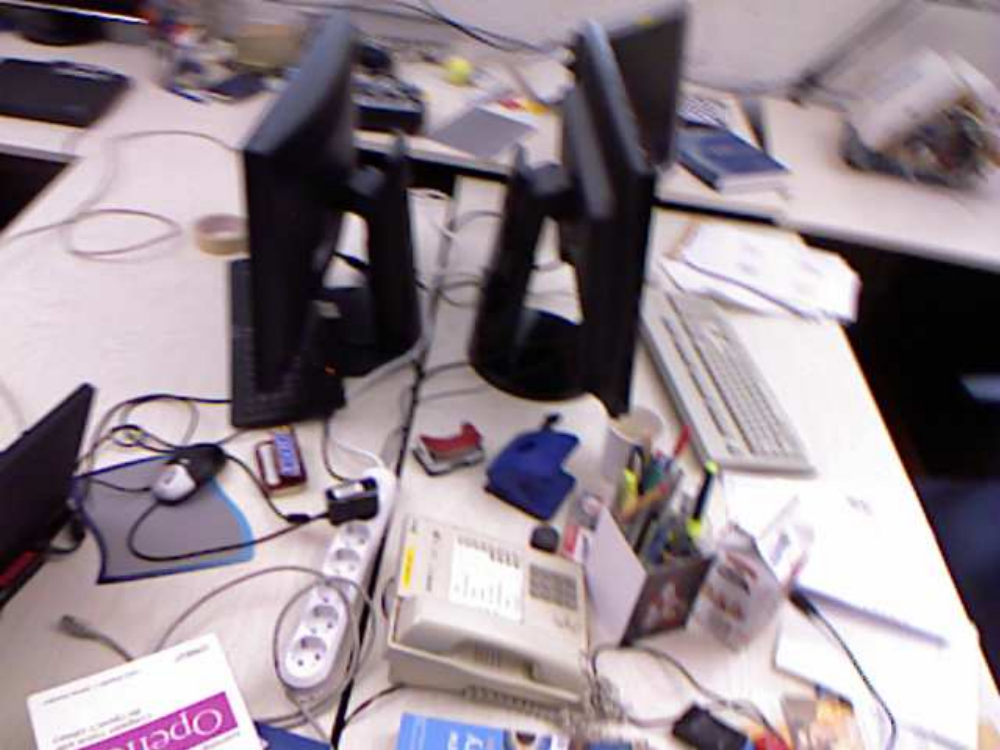}\label{exp:f3}}
\subfigure{\includegraphics[width=0.3\textwidth]
{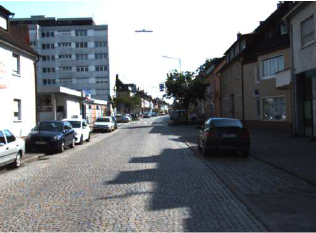}\label{exp:f5}} \\
\subfigure{\includegraphics[width=0.3\textwidth]
{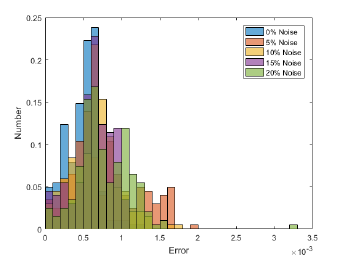}\label{exp:f2}}
\subfigure{\includegraphics[width=0.3\textwidth]
{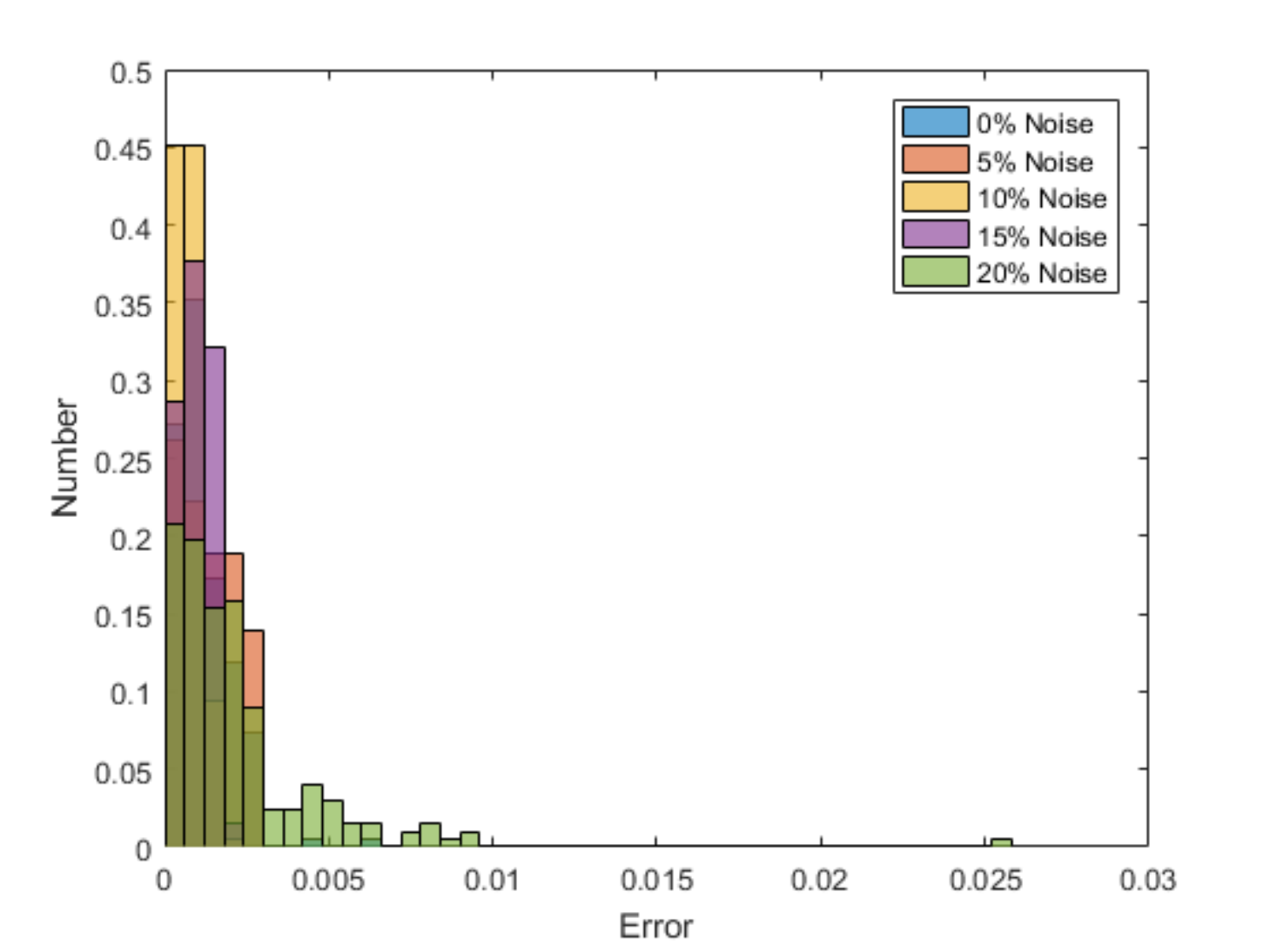}\label{exp:f4}}
\subfigure{\includegraphics[width=0.3\textwidth]{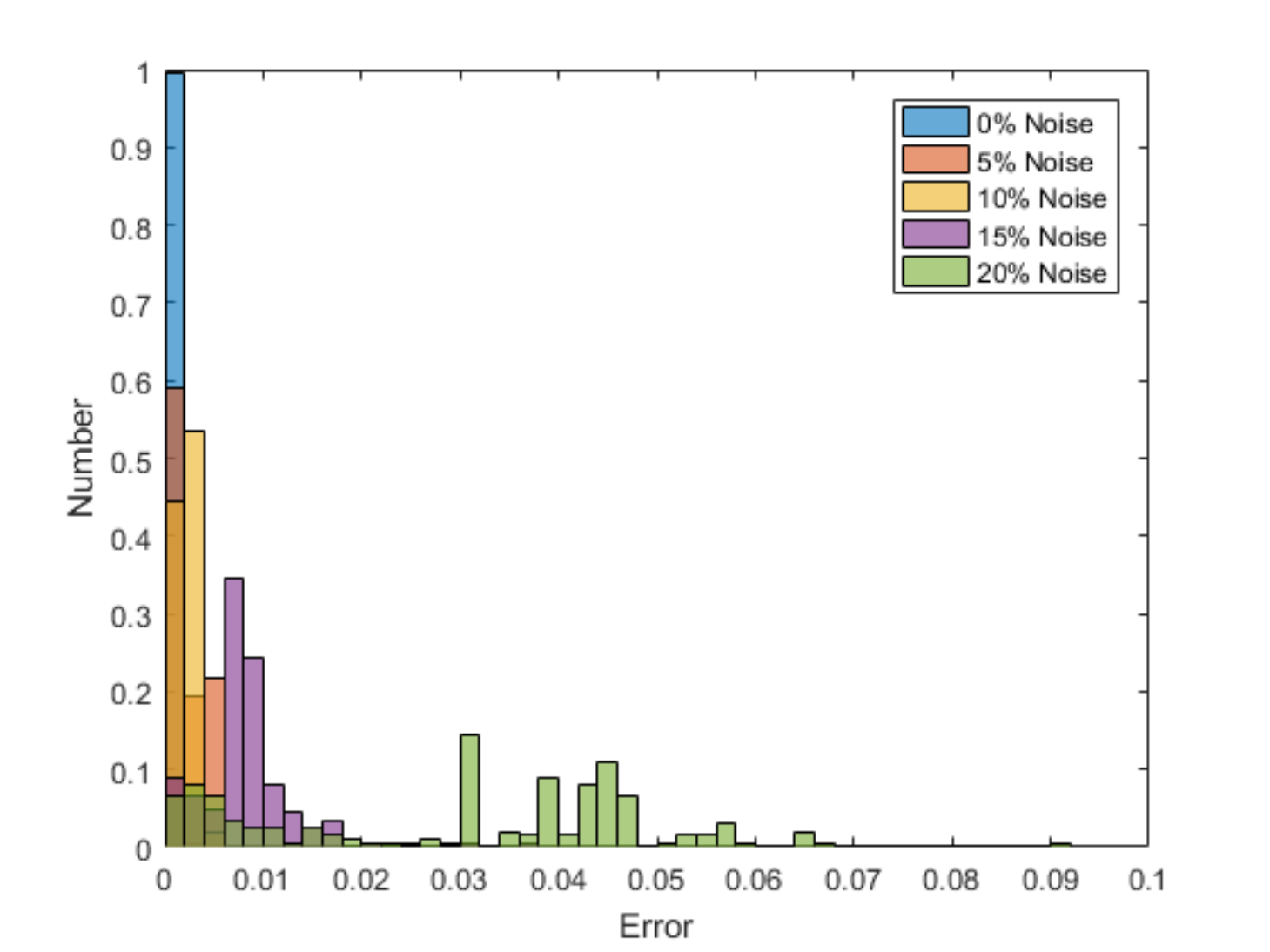}\label{exp:f6}}
\caption{The first row shows example images for each type of experiment. The second row shows the obtained distributions of position errors between the estimated result and ground truth for varying noise levels. For each pattern, we evaluate 5 different noise levels.}
\label{fig:Pattern}
\end{figure}

Three types of images are explored: logos, indoor environments (images taken from \cite{sturm2012benchmark}), and outdoor environments (images taken from \cite{Geiger2012CVPR}. We analyze three image for each category while each time add a varying amount of Gaussian noise to each individual image. Noise is added by adding a random per-pixel intensity disturbance of $0\%$ to $20\%$. A visualization of some patterns plus the distributions of obtained the position errors for the different noise levels are indicated in Figure \ref{fig:Pattern}.

As can be observed in Tables \ref{fig:table1} and \ref{fig:table2}, ORB-SLAM quickly fails as noise is added to the images, especially for the logo patterns which do not contain much texture. In contrast, the proposed curve-based optimization shows that, if a reasonably good initialization for poses is available, a high level of accuracy and robustness can be achieved for all analysed images.

\subsection{Evaluation on a full 3D dataset}

We evaluate the complete algorithm on the living room sequence of the ICML-NUIM benchmark sequence \cite{handa:etal:ICRA2014}. This synthetic dataset provides realistic images of a camera exploring an indoor environment. Both ground-truth information for poses and 3D model are available, thus permitting the evaluation of both accuracy of motion and quality of the reconstruction. As we use only the RGB channel of the dataset, the recovered scale of the estimation is in fact arbitrary. To properly evaluate the output trajectory, we therefore perform a 7 DoF alignment between the estimated trajectory and ground truth (i.e. a similarity transformation that identifies rotation, translation, and scale for an optimal alignment). The trajectory error is simply the distance between the recovered position and ground truth.

We run the pipeline ten times and compute the average rmse and median of the trajectory error for both ORB SLAM and our B\'ezier-spline based optimization. The results are indicated in Table \ref{fig:table3}. As can be observed, ORB-SLAM has a lower rmse error while using B\'ezier splines achieves a smaller median error. While this indicates generally better accuracy, the inferior performance in terms of the rmse error is attributed to a few occasions in the dataset where only few contours are observed, thus leading to no substantial improvement in the optimized pose. Note that, in order to further explore the potential of curve-based optimization, we also analysed the quality if more curves are initialized by also using the available depth channel (note however that depth is only used to initialize the curves, it does not constrain the curves during bundle adjustment anymore). The result is indicated in the last row of Table \ref{fig:table3}. It shows that if sufficient curves can be initialized, curve-based optimization is at last able to outperform state-of-the-art point-based approaches.

\begin{table}[h]
\caption{Average errors for ORB-SLAM \cite{mur2015orb} and our B\'ezier-spline based optimization.}
\label{fig:table3}
\begin{tabular}{p{0.22\linewidth}p{0.22\linewidth}p{0.22\linewidth}p{0.22\linewidth}|}
\hline
Algorithm & rmse(cm) & mean(cm) & median(cm)\\
\hline
ORB(Mono) & 3.82 & 3.41 & 3.02\\
Bezier & 4.45 & 3.68 & 2.96\\
Bezier(RGBD) & 4.11 & 3.07 & 2.37\\
\hline
\end{tabular}
\end{table}

\begin{figure}[t!]
\centering
\subfigure[]{\includegraphics[width=0.42\textwidth]
{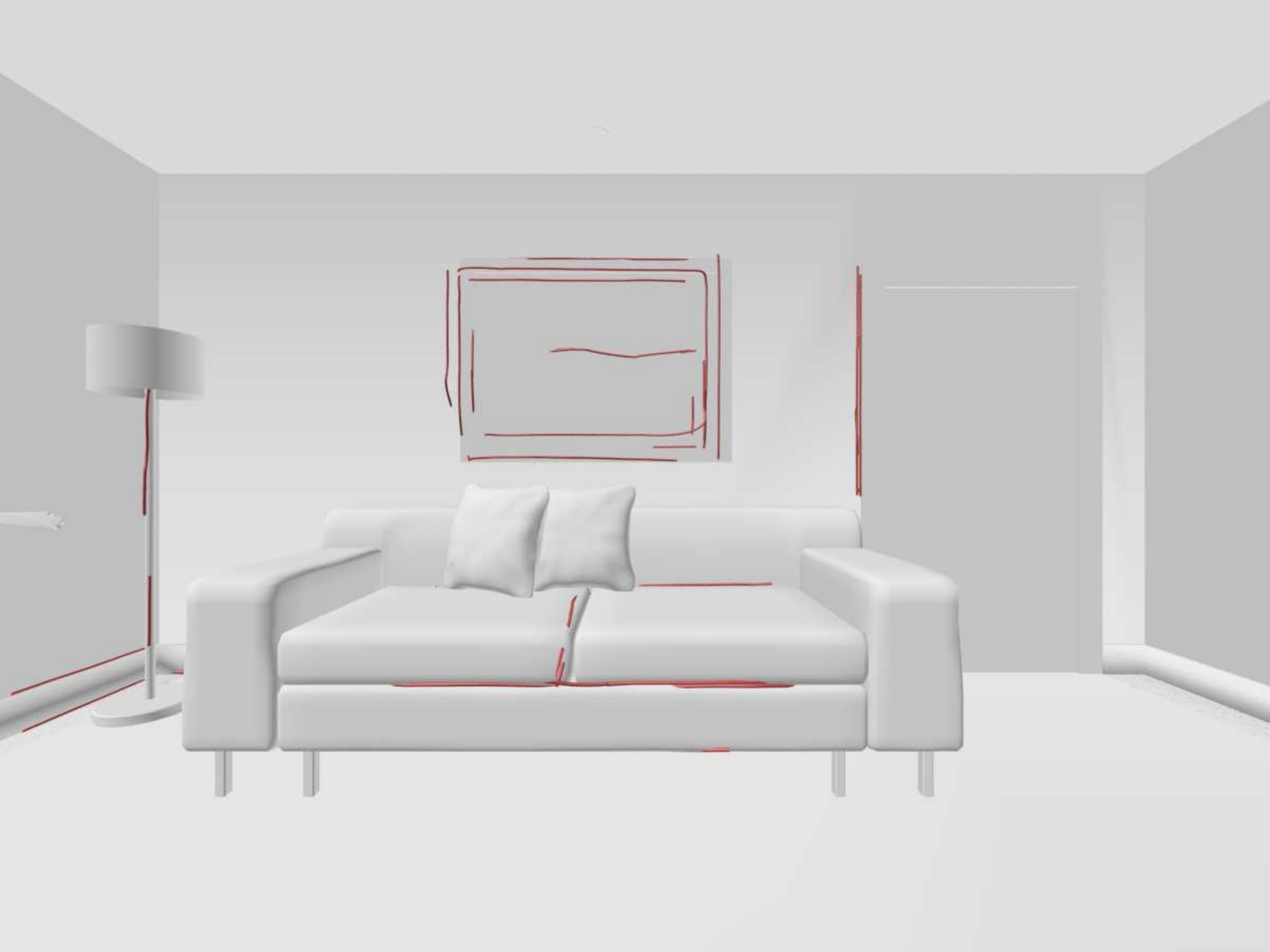}\label{eva:f1}}
\subfigure[]{\includegraphics[width=0.42\textwidth]
{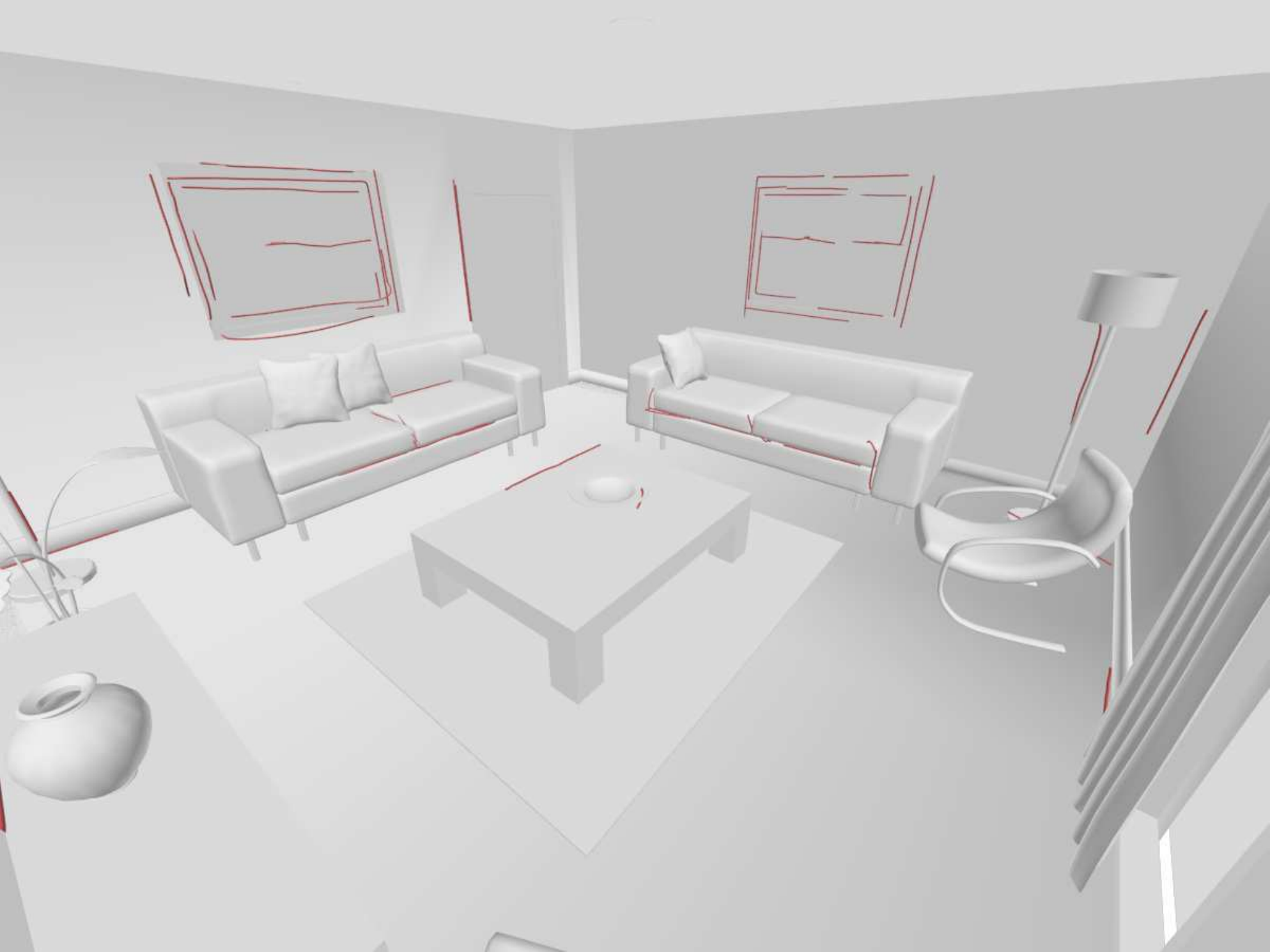}\label{eva:f2}}
\caption{Mapping results on the ICML-NUIM living room sequence \cite{handa:etal:ICRA2014}.}
\end{figure}

Since the ICML-NUIM dataset also provides 3D models of the environment, we can also visualize the quality of the mapping by overlaying some of our estimated curves onto the groundtruth CAD model of the environment. As illustrated in Figures \ref{eva:f1} and \ref{eva:f2}, the curves align well with real-world edges, and thus provide a visually appealing, more meaningful representation of the environment than sparse point-based approaches.
 
\section{Discussion}
\label{sec:discussion}

Our main novelty is a fully automatic structure from motion pipeline where a general, higher-order curve model has been successfully embedded into global bundle adjustment. This result stands in contrast with prior semi-dense visual SLAM pipelines, which alternate between tracking and mapping, and thus are unable to provide a globally consistent, jointly optimised result that explores all correlations between poses and structure. We employ polyb\'eziers, the geometric intuition of which proves great benefits during initialization and regularisation. Our work furthermore illustrates the importance of managing the correspondences between segments and frames, and the resulting graphical form of the optimisation problem. Introducing such correspondences also enables us to prevent the use of the more computationally demanding data-to-model registration paradigm. We present an evaluation on several synthetically generated datasets simulating the appearance of different environments and application scenarios. We demonstrate that it is indeed possible to improve on the accuracy provided by purely sparse methods, and return visually expressive, complete semi-dense models that are jointly optimized over all frames.


\newpage
\bibliographystyle{plainnat}
\bibliography{mybib}

\end{document}